\documentclass[
]{ceurart}

\usepackage{listings}
\usepackage[disable]{todonotes}
\usepackage{amsmath,amssymb}
\usepackage{tikz}
\usetikzlibrary{arrows.meta,positioning,shapes,fit,backgrounds}
\usepackage{xspace}
\newcommand{\probgen}{\textsc{ProbGen}\xspace}

\usepackage{numprint}

\npstyleenglish    %

\usepackage[capitalise,nameinlink]{cleveref}
\usepackage{algorithm}
\usepackage[noend]{algpseudocode}

\begin{document}

\copyrightyear{2025}
\copyrightclause{Copyright for this paper by its authors.
  Use permitted under Creative Commons License Attribution 4.0
  International (CC BY 4.0).}

%
%
\conference{}

\title{Quantifier Instantiations: To Mimic or To Revolt?}

\author{Jan Jakub\r uv}
[%
orcid=0000-0002-8848-5537,
url=https://people.ciirc.cvut.cz/~jakubja5/
]
\author{Mikol\'a\v s Janota}[%
orcid=0000-0003-3487-784X,
url=https://people.ciirc.cvut.cz/~janotmik
]
\address[]{Czech Technical University in Prague, Prague, Czech Republic}

\begin{abstract}
Quantified formulas pose a significant challenge for Satisfiability Modulo
Theories (SMT) solvers due to their inherent undecidability. Existing
instantiation techniques, such as e-matching, syntax-guided, model-based,
conflict-based, and enumerative methods, often complement each other. This
paper introduces a novel instantiation approach that dynamically learns from
these techniques during solving. By treating observed instantiations as samples
from a latent language, we use probabilistic context-free grammars to generate
new, similar terms. Our method not only mimics successful past instantiations
but also explores diversity by optionally inverting learned term probabilities,
aiming to balance exploitation and exploration in quantifier reasoning.
\end{abstract}
\

\begin{keywords}
  Satisfiability Modulo Theories \sep
  Quantifier Instantiation \sep
  Probabilistic Term Generation
\end{keywords}
\maketitle
\section{Introduction}
Solving formulas involving quantifiers is notoriously hard due to the
undecidabilty of the problem. Satisfiability Modulo Theories (SMT) solvers
tackle quantifiers by instantiating quantified variables by ground terms. A
series of techniques exists that attempt to find instantiations that will most
likely lead to a contradiction. Prominently, the techniques used in modern SMT solvers are
syntax-driven
(\emph{e-matching}~\cite{DetlefsNS05} or \emph{syntax-guided
instantiation}~\cite{niemetz-tacas21}), semantic-driven
(\emph{model-based}~\cite{ge-moura-cav09,reynolds-cade13}),
\emph{conflict-based}~\cite{reynolds-fmcad14}, and \textit{enumerative
instantiation}~\cite{janota2021fair,ReynoldsBF18}.
These techniques tend to give highly complementary results.

In this paper we propose a quantifier instantiation technique that seeks
inspiration in the other techniques on the fly---while solving the formula. The
idea is as follows. Observe which instantiations were made by other techniques
and then, attempt to make similar ones. Now, how to find similar instantiations? We
look for inspiration in \emph{probabilistic context free grammars}~\cite{charniak97}.
We imagine the instantiations made so far as a language, where each ground term
is a sentence in the language. Since we do not have an explicit definition of
the language, we take our observations as samples from it.

\begin{figure}[ht]
  \centering

\begin{tikzpicture}[
    box/.style={draw, thick, minimum width=2.5cm, minimum height=1em, align=center},
    circle node/.style={draw, circle, fill=white, inner sep=2pt},
    arrow/.style={->, >=Stealth, thick},
    node distance=2em,
    yscale=.4
]

\node[box,fill=green!20] (module1) at (-3,0.5) {Inst.\ module 1};
\node[box,fill=green!20] (module2) at (-3,-1) {Inst.\ module 2};
\node (dots) at (-3,-2.5) {$\ldots$};
\node[box, align=center,fill=red!20] (probgen) at (0,-5) {\textsc{Probgen}};
\node[box, align=center,fill=green!20] (solver) at (3,0.5) {Ground Solver};

\node[circle node] (c1) at (0,.5) {};
\draw[arrow] (c1) -- (probgen);
\node[circle node] (c2) at (0,-1) {};
\node[circle node] (c3) at (0,-2.5) {};

\draw[arrow] (module1) -- (c1);
\draw[arrow] (module2) -- (c2);
\draw[arrow] (dots) -- (c3);
\draw[arrow] (c1) -- (solver);
\draw[arrow] (c2) -| (solver.south west);
\draw[arrow] (c3) -| (solver);
\draw[arrow] (probgen) -| (solver.south east);

\end{tikzpicture}
   \caption{Schematic depiction of the overall idea, where Probgen is
  our contribution}\label{fig:idea}
\end{figure}
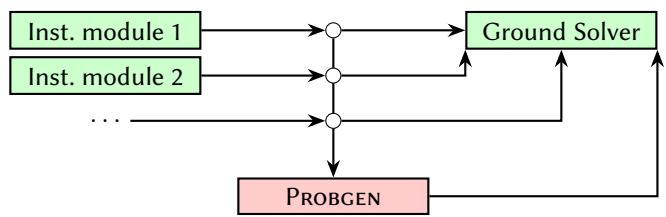

\autoref{fig:idea} illustrates the overall idea of the approach.  Instantiation
is made by several instantiation modules, e.g.\ e-matching. Each instantiation
module produces some instantiations that are sent to the ground solver. In our
approach, these are also intercepted and collected in the \probgen generator,
which also produces its own instantiations sent to the ground solver.

The idea of probabilistic grammars can be simplified even further. Let's say
that the constant $c$ appears frequently under the function symbol $f$, then
$f(c)$ should also be a likely term in our language. However, this suggests a
different question. If the new terms are generated according to the ones that
were already used in the past, will they be useful? Should we not rather do the
opposite, i.e., \emph{generate different terms}?  Effectively, this means
\emph{inverting} the probabilities when generating new terms for instantiation.

\section{Preliminaries}\label{sec:preliminaries}

Basic understanding of SMT~\cite{barrettSATHandbook} is assumed. In special
cases, such as linear arithmetic, decision procedures exist for theories with
quantifiers~\cite{Weispfenning:88,Tarski:51,Davenport:88,bjorner-lpar15}. 
In general, however, presence of quantifiers can easily make the problem
undecidable and in SMT, quantifier instantiations are used to tackle this
problem.
A quantified subformula $\forall\bar x.\phi$, with a vectors of variables $\bar
x$ is abstracted as a proposition $Q$ and instantiations are realized as
implications of the form $Q\rightarrow\phi[\bar x\mapsto\bar t]$, where $t$  is
a vector of ground terms. This implication is called a \emph{lemma}. %
A plethora of instantiation techniques appear in the literature. These
typically exhibit complementary behavior and do not interact directly with one
another, although indirect interaction through a shared solver state can occur.

Probabilistic grammars~\cite{jurafsky2023slp,charniak97}, extend traditional
formal grammars by associating probabilities with their production rules. This
probabilistic framework allows for modeling uncertainty and variation in natural
language and other structured data. In a probabilistic context-free grammar
(PCFG), each rule of the form $A \rightarrow \alpha$ is assigned a probability
$P(A \rightarrow \alpha)$, such that the sum of probabilities of all rules with
the same left-hand nonterminal $A$ is 1. These models are particularly useful in
computational linguistics and natural language processing tasks, where ambiguity
and multiple possible interpretations are common. PCFG have been used in
autoformalization tasks~\cite{kaliszyk2017}.

\section{Term Generation}\label{sec:term_generation} %

Full-fledged learning of PCFG is likely to be to expensive for our purposes and
we take a Markov-like approach instead. New ground terms are generated
recursively, as a tree, and the probability of a symbol $s$ being generated in a
node $n$ is determined by the frequency of $s$ in the instantiations so far.

\begin{algorithm}[t]
\caption{Make Term}\label{alg:mkterm}
\begin{algorithmic}[1]
\item[\textbf{Input:}] $\tau$ - range type, $d$ - term depth, initially $0$
\item[\textbf{Output:}] term of type $\tau$
\item[\textbf{Using:}] $\textproc{Symbols}(\tau)$ - symbols constructing terms of type $\tau$
\item[\hspace{2.9em}] $\textproc{Depth}$ - maximum allowed term depth
\item[\hspace{2.9em}] $\textproc{Pick}(S)$ - pick one symbol out of $S$ according to the selection method
\item[\hspace{2.9em}] $\textproc{Arity}(s)$ - the arity of symbols $s$
\Function{MakeTerm}{$\tau, d=0$}
    \State $S \gets \textproc{Symbols}(\tau)$
    \If{$d \geq \textproc{Depth}$}
        \State $S \gets \{s \in S \mid s \text{ is a constant}\}$
    \EndIf
    \State $s \gets \textproc{Pick}(S)$
    \If{$s$ is a constant}
        \State \Return $s$
    \Else
        \ForAll{$0 < i \leq \textproc{Arity}(s)$}
            \State $t_i \gets \textproc{MakeTerm}(\tau_i, d+1)$ 
            \textbf{ where }  $\tau_i \text{ is the type of the $i$-th argument of $s$} $
        \EndFor
        \State \Return $s(t_1, \ldots, t_n)$
    \EndIf
\EndFunction
\end{algorithmic}
\end{algorithm}

Terms are generated recursively in a straightforward fashion with fixed maximum
depth using function \textproc{MakeTerm} in \cref{alg:mkterm}. 
The function $\textproc{MakeTerm}(\tau)$ generates a term of type $\tau$ by
selecting a term head symbols and recursively generating argument terms, possibly
resorting to constants to limit the depth of generated terms.
The head symbol is always picked from the set $\textproc{Symbols}(\tau)$ which
contains all symbols of type $\tau$ encountered in the terms generated by other instantiation modules.
The maximum allowed term depth is controlled by parameter \textproc{Depth}.
With $\textproc{Depth}=0$, the algorithm generates constants only.

The key point of the algorithm is the selection of the term head symbol using
the function \textproc{Pick}. Naturally, there is a lot of flexibility in this
choice, and that is where our statistical approach comes into play. We
implement three different selection methods. The first one, denoted
\textproc{random}, selects a symbol uniformly at random, independent of any
symbol statistics. The next method, \textproc{weights}, maintains a statistic
for each symbol $s$, counting the number of times $s$ occurs in terms generated
by other instantiation techniques. These counts are interpreted as
probabilistic weights, and the head symbol is then sampled from a categorical
distribution derived from them. In practice, this means treating the weights as
proportional intervals on a number line, generating a random number in the
range from $0$ to the total weight sum, and selecting the symbol corresponding
to the interval in which the number falls.
The third and final selection method we experiment with is \textproc{paths},
which extends \textproc{weights} by taking term positions into account. It
maintains separate weight vectors for each term position, where a position is
determined by the list of symbols occurring above the term in the syntax tree.
For example, the position of both $a$ and $b$ in $f(g(a,b))$ is given by the
path $(f, g)$.
Since the weight vectors for \textproc{paths} can be quite sparse and contain
only few symbols of the given type, we additionally consider all other
compatible symbols with the default weight $1$ at each position.

The above symbol selection methods \textproc{weights} and \textproc{paths}
generates terms mimicking terms generated by other instantiation modules. In
order to introduce diversity and generate different terms, we introduce
additional parameter \textproc{Flip}. Its value is the probability under which
the weights should be inverted, setting all weights to $1/w$ instead of $w$.
This probability is applied in every call to function \textproc{Pick}. Hence
different calls to \textproc{Pick} within one call to \textproc{MakeTerm} can
work with different weights.

The final parameter we introduce to our probabilistic instantiation module
concerns when the module is activated. In cvc5, several \emph{effort} levels
are defined for instantiation modules. These modules are tried sequentially,
with the effort level increasing if no lemma is produced at the current level.
The currently implemented effort levels in cvc5, in order of increasing effort,
are \emph{conflict}, \emph{standard}, \emph{model}, and \emph{last call}.
Conflict-based instantiation modules are tried first (\emph{conflict}),
followed by heuristic instantiation modules (\emph{standard}, e.g.,
e-matching), then model-based techniques (\emph{model}), and finally full
effort instantiation (\emph{last call}) are launched.

We follow the behavior of the enumerative instantiation modules and introduce a
parameter \textproc{Effort}, which supports two values: \textproc{lastcall} and
\textproc{interleave}. With \textproc{lastcall}, our probabilistic
instantiation module is executed only at the last call effort level, that is,
when no module has produced a lemma at a lower effort. With
\textproc{interleave}, instantiations are performed already at the standard
effort level, meaning the module runs more frequently than with
\textproc{lastcall}.

\section{Experimental Evaluation}\label{sec:experiments}

We evaluate%
\footnote{On a server with two AMD EPYC 7513 32-Core processors @ 3680 MHz and
with 514 GB RAM.}
our approach on a benchmark from SMT-LIB~\cite{BarFT-RR-17,BarFT-SMTLIB},
namely on $8,024$ problems from the \texttt{2019-Preiner} directory of the UFNIA
logic. This benchmark was chosen based on a preliminary experiment on a larger
subset of SMT-LIB since our methods seemed to perform well therein.
We perform an extensive grid search for various parameters of the probabilistic
instantiation module, namely all of the following combinations. 
\begin{displaymath}
   \begin{array}{rcl}
      \textproc{Effort}   & \in & \{ \textproc{lastcall}, \textproc{interleave} \} \\
      \textproc{Pick}     & \in & \{ \textproc{random}, \textproc{weights}, \textproc{paths} \} \\
      \textproc{Depth}    & \in & \{ 0,1,2,3,4 \} \\
      \textproc{Flip}     & \in & \{ 0.0,0.2,0.5,0.8,1.0 \} \\
   \end{array}
\end{displaymath}
Since the parameter \textproc{Flip} is applicable only when \textproc{Pick} is either 
\textproc{weights} or \textproc{paths}, we obtain altogether $110$ different strategies.
We evaluate all of them with a $10$ second time limit per strategy and problem.
Our probabilistic instantiation module is launched together with the default
cvc5 setting which uses e-matching and conflict-based quantifier instantiation
(cbqi).
The instantiation terms introduced by these two default modules are intercepted
and used to collect symbol statistics utilized by our \textproc{weights} and
\textproc{paths} symbol selection functions.
The set of asserted quantifiers is randomly shuffled in each round.
This is because we terminate the instantiation round after a first successfully
generated lemma.

We always generate $20$ terms for each encountered type and remove duplicates
if necessary.
Within one instantiation round, only one set of terms is generated for every
encountered type, that is, for every type of a $\forall$-bound variable from 
asserted quantifiers.
This gives us a non-zero probability that variables of the same type will be
instantiated by the same term.
A new set of terms is generated in the next instantiation round.

\begin{figure*}[t]
\footnotesize
\begin{center}
\def\arraystretch{1.2}%
\setlength\tabcolsep{0.25em}%
\begin{tabular}{c||lc|lc|lc||lc|lc|lc}
   \textproc{Effort} & 
   \multicolumn{6}{c||}{\textproc{lastcall}} &
   \multicolumn{6}{c}{\textproc{interleave}}
\\\hline
   \textproc{Pick} & 
   \multicolumn{2}{c|}{\textproc{random}} & 
   \multicolumn{2}{c|}{\textproc{weights}} & 
   \multicolumn{2}{c||}{\textproc{paths}} &
   \multicolumn{2}{c|}{\textproc{random}} & 
   \multicolumn{2}{c|}{\textproc{weights}} & 
   \multicolumn{2}{c}{\textproc{paths}} 
\\\hline
   \textproc{Depth} & 
   \emph{total} & \emph{ref (+/-)} &
   \emph{total} & \emph{ref (+/-)} &
   \emph{total} & \emph{ref (+/-)} &
   \emph{total} & \emph{ref (+/-)} &
   \emph{total} & \emph{ref (+/-)} &
   \emph{total} & \emph{ref (+/-)} 
\\\hline
   $0$   &    $\mathbf{3523}$   &   +$\mathbf{499}$/-$53$    &    $\mathbf{3497}$   &   +$\mathbf{477}$/-$\mathbf{57}$    &    $\mathbf{3445}$   &   +$\mathbf{427}$/-$59$    &    $\mathbf{\underline{3575}}$   &   +$\mathbf{\underline{571}}$/-$\mathbf{73}$    &    $\mathbf{3570}$   &   +$\mathbf{\underline{571}}$/-$\mathbf{78}$    &    $\mathbf{3498}$   &   +$\mathbf{497}$/-$\mathbf{76}$    \\
$1$   &    $3293$   &   +$270$/-$54$    &    $3245$   &   +$230$/-$62$    &    $3273$   &   +$247$/-$51$    &    $3309$   &   +$321$/-$89$    &    $3284$   &   +$300$/-$93$    &    $3327$   &   +$349$/-$99$    \\
$2$   &    $3221$   &   +$195$/-$51$    &    $3231$   &   +$215$/-$61$    &    $3220$   &   +$190$/-$47$    &    $3242$   &   +$259$/-$94$    &    $3239$   &   +$256$/-$94$    &    $3211$   &   +$235$/-$101$    \\
$3$   &    $3220$   &   +$193$/-$\mathbf{50}$    &    $3192$   &   +$182$/-$67$    &    $3219$   &   +$188$/-$\mathbf{\underline{46}}$    &    $3276$   &   +$290$/-$91$    &    $3171$   &   +$211$/-$117$    &    $3230$   &   +$256$/-$103$    \\
$4$   &    $3212$   &   +$197$/-$62$    &    $3181$   &   +$169$/-$65$    &    $3215$   &   +$184$/-$\mathbf{\underline{46}}$    &    $3236$   &   +$258$/-$99$    &    $3161$   &   +$206$/-$122$    &    $3228$   &   +$255$/-$104$    \\
\end{tabular}
\end{center}
\normalsize
\caption{Grid evaluation of strategies with different \textproc{Depth} values (with fixed $\textproc{Flip}=0.0$).}
\label{fig:depths}
\end{figure*}

The results for different values of \textproc{Depth} with \textproc{Flip} fixed to $0.0$ are depicted in 
\autoref{fig:depths}.
This allows us to evaluate the effect of various term depths.
For each strategy we present three numbers: (1) the total number of solved
problems (column \emph{total}), and (2-3) the number of solutions gained ($+$)
and lost ($-$) on the baseline reference strategy (columns \emph{ref}). As the
baseline strategy we consider cvc5 with the default option setting. 
\textbf{The reference strategy solves \numprint{3077} problems.}
To ease orientation, the best value in each column is \textbf{highlighted} and
the best values within the table are additionally \underline{underlined}.

\noindent We can draw several observations from \autoref{fig:depths}:
\begin{enumerate}
\item All strategies significantly outperform the reference strategy, which
   solves \numprint{3077} problems.
\item In every case, increasing the maximum term depth leads to a decline in
   performance. The best results are achieved when instantiating with
   constants only ($\textproc{Depth}=0$), likely due to the reduced
   complexity of the term space at lower depths.
\item Strategies using the \textproc{interleave} effort, where our
   instantiation module is invoked more frequently, tend to perform better
   than those using \textproc{lastcall}.
   This suggests that there is a potential advantage in our approach.
\item Each strategy both solves some problems that the reference does not
   (\emph{ref+}) and fails on some that the reference does solve
   (\emph{ref-}). The \textproc{lastcall} strategies lose fewer solutions
   because the probabilistic instantiation is applied less frequently,
   resulting in behavior closer to the reference. The fewest losses occur
   with the \textproc{paths} variants at higher term depths, where the
   generated terms more closely resemble those of the reference strategy.
\item There appears to be no significant advantage of probabilistic
   generation over random term generation in this case. Since the depth is
   fixed to $0$, the \textproc{paths} variant does not fully benefit from using
   different weight vectors for different positions. Moreover, the
   \textproc{weights} variant instantiates using all constants from the other
   instantiation modules, guided by probabilities based on their occurrence
   counts. In contrast, \textproc{paths} in this case only tracks statistics
   for constants that appear at the top level while all other constants are
   assigned a default weight of $1$.
\end{enumerate}

\begin{figure*}[t]
\footnotesize
\begin{center}
\def\arraystretch{1.3}%
\setlength\tabcolsep{0.9em}%
\begin{tabular}{c||lc|lc||lc|lc}
   \textproc{Effort} & 
   \multicolumn{4}{c||}{\textproc{lastcall}} &
   \multicolumn{4}{c}{\textproc{interleave}}
\\\hline
   \textproc{Pick} & 
   \multicolumn{2}{c|}{\textproc{weights}} & 
   \multicolumn{2}{c||}{\textproc{paths}} &
   \multicolumn{2}{c|}{\textproc{weights}} & 
   \multicolumn{2}{c}{\textproc{paths}} 
\\\hline
   \textproc{Flip} & 
   \emph{total} & \emph{ref (+/-)} &
   \emph{total} & \emph{ref (+/-)} &
   \emph{total} & \emph{ref (+/-)} &
   \emph{total} & \emph{ref (+/-)} 
\\\hline
$0.0$   &    $3497$   &   +$477$/-$57$    &    $3445$   &   +$427$/-$59$    &    $3570$   &   +$571$/-$78$    &    $3498$   &   +$497$/-$76$    \\                       
$0.2$   &    $\mathbf{3509}$   &   +$\mathbf{500}$/-$68$    &    $3479$   &   +$463$/-$61$    &    $3579$   &   +$578$/-$76$    &    $3510$   &   +$504$/-$\mathbf{71}$    \\                               
$0.5$   &    $3474$   &   +$475$/-$78$    &    $3488$   &   +$467$/-$56$    &    {\normalsize$\mathbf{\underline{3613}}$}   &   +$\mathbf{\underline{612}}$/-$76$    &    $3510$   &   +$508$/-$75$    \\                               
$0.8$   &    $3507$   &   +$493$/-$63$    &    $\mathbf{3515}$   &   +$\mathbf{496}$/-$58$    &    $3579$   &   +$575$/-$\mathbf{73}$    &    $3517$   &   +$516$/-$76$    \\                               
$1.0$   &    $3335$   &   +$314$/-$\mathbf{56}$    &    $3495$   &   +$473$/-$\mathbf{\underline{55}}$    &    $3429$   &   +$426$/-$74$    &    $\mathbf{3532}$   &   +$\mathbf{530}$/-$75$    \\
\end{tabular}
\end{center}
\normalsize
\caption{Grid evaluation of strategies with different \textproc{Flip} values (with fixed $\textproc{Depth}=0$).}
\label{fig:flips}
\end{figure*}

In the next \autoref{fig:flips} we fix \textproc{Depth} to $0$ and we evaluate
various values for \textproc{Flip}.
Higher values force our module to generate terms more different than term used
by default instantiation modules.
The table format is the same as in \autoref{fig:depths} with the exception that
the \textproc{random} case is omitted since the \textproc{Flip} parameter does
not effect it.
The line for $\textproc{Flip} = 0$ has the same values as in the line
$\textproc{Depth} = 0$ in the previous figure.

The best results are obtained with the \textproc{weights} variant using the
\textproc{interleave} effort and $\textproc{Flip} = 0.5$. This strategy solves
\numprint{3613} problems and produces \numprint{612} new solutions compared to
the baseline. Overall, we observe that better performance is achieved when
$\textproc{Flip}$ is strictly greater than $0.0$, suggesting that occasionally
generating complementary terms is beneficial. By tuning the \textproc{Flip}
parameter, we outperform the random term generation results shown in
\autoref{fig:depths}.

\begin{figure*}[t]
\begin{center}
\def\arraystretch{1.05}%
\setlength\tabcolsep{0.9em}%
\begin{tabular}{lrlr||rrl|l}
   \multicolumn{4}{c||}{\emph{strategy parameters}} &
   \multicolumn{4}{c}{\emph{performance in greedy cover}}
\\
   \textproc{Effort} & \textproc{Depth} & \textproc{Pick} & \textproc{Flip} & \emph{solves} & \emph{+new} & \emph{adds} & $=$ \emph{total} 
\\\hline
\textproc{interleave} & $0$  & \textproc{weights}  & $0.5$          & $3613  $ & $+3613  $ & $-       $ & $=3613  $ \\
\textproc{standard  } & $0$  & \textproc{weights}  & $0.2$          & $3509  $ & $+141   $ & $+3.90\% $ & $=3754  $ \\
\textproc{standard  } & $0$  & \textproc{weights}  & $0.0$          & $3497  $ & $+82    $ & $+2.18\% $ & $=3836  $ \\
\textproc{standard  } & $0$  & \textproc{random }  & $-$            & $3523  $ & $+52    $ & $+1.36\% $ & $=3888  $ \\
\textproc{interleave} & $0$  & \textproc{paths  }  & $0.0$          & $3498  $ & $+42    $ & $+1.08\% $ & $=3930  $ \\
\textproc{standard  } & $0$  & \textproc{paths  }  & $0.2$          & $3479  $ & $+29    $ & $+0.74\% $ & $=3959  $ \\
\textproc{interleave} & $0$  & \textproc{weights}  & $0.8$          & $3579  $ & $+24    $ & $+0.61\% $ & $=3983  $ \\
\textproc{standard  } & $0$  & \textproc{weights}  & $0.8$          & $3507  $ & $+21    $ & $+0.53\% $ & $=4004  $ \\
\textproc{interleave} & $0$  & \textproc{weights}  & $0.0$          & $3570  $ & $+17    $ & $+0.42\% $ & $=4021  $ \\
\textproc{interleave} & $0$  & \textproc{random }  & $-$            & $3575  $ & $+14    $ & $+0.35\% $ & $=4035  $ \\
\textproc{interleave} & $0$  & \textproc{paths  }  & $0.5$          & $3510  $ & $+12    $ & $+0.30\% $ & $=4047  $ \\
\textproc{standard  } & $3$  & \textproc{random }  & $-$            & $3220  $ & $+11    $ & $+0.27\% $ & $=4058  $ \\
\textproc{interleave} & $0$  & \textproc{weights}  & $0.2$          & $3579  $ & $+10    $ & $+0.25\% $ & $=4068  $ \\
\textproc{interleave} & $0$  & \textproc{weights}  & $1.0$          & $3429  $ & $+9     $ & $+0.22\% $ & $=4077  $ \\
\textproc{standard  } & $0$  & \textproc{paths  }  & $1.0$          & $3495  $ & $+8     $ & $+0.20\% $ & $=4085  $ \\
\textproc{standard  } & $0$  & \textproc{weights}  & $0.5$          & $3474  $ & $+7     $ & $+0.17\% $ & $=4092  $ \\
\textproc{standard  } & $1$  & \textproc{paths  }  & $0.5$          & $3124  $ & $+7     $ & $+0.17\% $ & $=4099  $ \\
\textproc{standard  } & $0$  & \textproc{paths  }  & $0.0$          & $3445  $ & $+6     $ & $+0.15\% $ & $=4105  $ \\
\textproc{standard  } & $1$  & \textproc{random }  & $-$            & $3293  $ & $+6     $ & $+0.15\% $ & $=4111  $ \\
\textproc{interleave} & $0$  & \textproc{paths  }  & $0.8$          & $3517  $ & $+5     $ & $+0.12\% $ & $=4116  $ \\
\end{tabular}
\end{center}
\normalsize
\caption{Greedy cover from the evaluated strategies helps to measure complementarity.}
\label{fig:greedy}
\end{figure*}

The final experiment in this work aims to evaluate the mutual complementarity
of the tested strategies. A greedy cover sequence is constructed from all
evaluated strategies. The sequence is initialized by selecting the most
effective individual strategy, and the problems it solves are marked. At each
subsequent step, the strategy that solves the largest number of remaining
unsolved problems is selected. This process is repeated iteratively. In this
way, the greedy cover reveals strategies that complement one another.

The first 20 strategies in the greedy cover are presented in
\autoref{fig:greedy}. The first four columns specify the strategy parameters,
while the next four columns summarize their performance within the greedy
cover. The column \emph{solves} shows the number of problems each strategy
solves individually. The remaining columns reflect each strategy's contribution
to the overall portfolio performance. The column \emph{+new} indicates how many
additional problems the strategy contributes to the portfolio. The column
\emph{total} gives the cumulative number of problems solved by the portfolio up
to that point, computed as the sum of \emph{+new} and the previous row's
\emph{total}. Finally, the column \emph{adds} expresses \emph{+new} as a
percentage of the current portfolio size.

We observe that the two most complementary strategies switch the effort
mode from \textproc{interleave} to \textproc{standard} and use different values
for \textproc{Flip}. Since the second strategy contributes 3.90\% to the
portfolio, this indicates a meaningful level of complementarity among the
strategies. Furthermore, we observe that the majority of strategies use a term
depth of $0$, meaning they instantiate using constants only. This suggests that
the strategies not shown in \autoref{fig:depths} and
\autoref{fig:flips}, specifically those with $\textproc{Depth} > 0$ and
$\textproc{Flip} > 0.0$, do not yield any significant improvement.

\section{Conclusions and Future Work}\label{sec:conclusions}

We have presented preliminary experiments on the probabilistic generation of
instantiation terms for quantified formulas in cvc5. The results indicate
potential in our approach, as we were able to improve upon the reference
baseline strategy. The best performance was achieved by strategies that
generate terms complementary to those produced by other instantiation
techniques. This highlights the importance of diversity in instantiation
strategies and suggests that probabilistic generation can effectively fill gaps
left by more deterministic methods. Leveraging this complementarity may be key
to further improving solver performance on quantified benchmarks. In contrast,
guided generation of more complex or deeper terms has not yet proven to be
effective.

Future work will focus on improving the guided generation of complex terms.
Given the exponential growth of the instantiation term space, an effective
reduction of the search space is essential. One possible direction is to employ
a learning-based approach to guide term generation. We would like to refine the
interaction between probabilistic instantiation and existing instantiation
techniques, exploring how to better coordinate or prioritize between them
during solving. 
Additionally, we plan to investigate adaptive mechanisms that
adjust term generation parameters dynamically based on solver progress or
formula structure. Another direction is to explore richer probabilistic
models, beyond simple frequency-based grammars, that can better capture
structural patterns in useful instantiations. 
Finally, we intend to evaluate our approach on broader and more diverse
benchmarks to better understand its strengths and limitations in different
theory combinations.

\section*{Acknowledgements}
 This work is supported by the Czech MEYS under the ERC~CZ project no.~LL1902
 \emph{POSTMAN}, and by the European Union under the project \emph{ROBOPROX}
 (reg.~no.~CZ.02.01.01/00/22\_008/0004590), and by the \emph{RICAIP} project
 that has received funding from the European Union's Horizon~2020 research and
 innovation programme under grant agreement No~857306.

\bibliography{refs}
\end{document}